\documentclass[11pt,a4paper]{article}
\usepackage[hyperref]{emnlp2018}
\usepackage{times}
\usepackage{latexsym}
\usepackage{bm}
\usepackage{amssymb, amsmath}
\usepackage{algorithm}
\usepackage[noend]{algpseudocode}
\usepackage{graphicx}  
\usepackage{subcaption}

\usepackage{url}

{\raise0.5ex\hbox{$#1$}\! \left/ \! \lower0.5ex\hbox{$#2$}\right.}

\aclfinalcopy 

\newcommand*\myat{{\fontfamily{ptm}\selectfont\small @}}

\title{Speeding Up Neural Machine Translation Decoding by Cube Pruning}

\def\first{$^1$}
\def\second{$^2$}
\def\third{$^3$}
\def\fourth{$^4$}
\def\fifth{$^5$}
\def\comma{$^,$}
\author{Wen Zhang\first\comma\second ~~~ Liang Huang\third\comma\fourth ~~~ Yang Feng\first\comma\second ~~~ Lei Shen\first\comma\second ~~~ Qun Liu\fifth\comma\first
\\
\\
{ \first {Key Laboratory of Intelligent Information Processing}} \\ Institute of Computing Technology, Chinese Academy of Sciences (ICT/CAS)   \\
{ \second {University of Chinese Academy of Sciences, Beijing, China}} \\
{\tt \small \{\href{mailto:zhangwen@ict.ac.cn}{zhangwen},\href{mailto:fengyang@ict.ac.cn}{fengyang},\href{mailto:shenlei17z@ict.ac.cn}{shenlei17z}\}\myat ict.ac.cn}	\\[0.15cm]
{ \third {Oregon State University, Corvallis, OR, USA}} \quad
{ \fourth {Baidu Research, Sunnyvale, CA, USA}}\\[-0.05cm]
\href{mailto:liang.huang.sh@gmail.com}{{\tt\small liang.huang.sh\myat gmail.com}}\\[0.15cm]
{ \fifth {Huawei Noah's Ark Lab, Hong Kong, China}}\\[-0.05cm]
\href{mailto:qun.liu@huawei.com}{{\tt\small qun.liu\myat huawei.com}} \\
}

\date{}

\begin{document}

\maketitle

\begin{abstract}
Although neural machine translation has achieved promising results, it suffers from slow translation speed. The direct consequence is that a trade-off has to be made between translation quality and speed, thus its performance can not come into full play. We apply cube pruning, a popular technique to speed up dynamic programming, into neural machine translation to speed up the translation. To construct the equivalence class, similar target hidden states are combined, leading to less RNN expansion operations on the target side and less $\mathrm{softmax}$ operations over the large target vocabulary. The experiments show that, at the same or even better translation quality, our method can translate faster compared with naive beam search by $3.3\times$ on GPUs and $3.5\times$ on CPUs.

\end{abstract}

\section{Introduction}
\label{sec1}
Neural machine translation (NMT) has shown promising results and drawn more attention recently \cite{kalchbrenner2013recurrent,cho2014learning,bahdanau2014neural,gehring2017cnn,gehring2017convolutional,vaswani2017attention}. A widely used architecture is the attention-based encoder-decoder framework \cite{cho2014learning,bahdanau2014neural} which assumes there is a common semantic space between the source and target language pairs. The encoder encodes the source sentence to a representation in the common space with the recurrent neural network (RNN) \cite{hochreiter1997long} and the decoder decodes this representation to generate the target sentence word by word. To generate a target word, a probability distribution over the target vocabulary is drawn based on the attention over the entire source sequence and the target information rolled by another RNN. At the training time, the decoder is forced to generate the ground truth sentence, while at inference, it needs to employ the beam search algorithm to search through a constrained space due to the huge search space.

Even with beam search, NMT still suffers from slow translation speed, especially when it works not on GPUs, but on CPUs, which are more common practice. The first reason for the inefficiency is that the generation of each target word requires extensive computation to go through all the source words to calculate the attention. Worse still, due to the recurrence of RNNs, target words can only be generated {\em sequentially} rather than in parallel. The second reason is that large vocabulary on target side is employed to avoid unknown words (UNKs), which leads to a large number of normalization factors for the $\mathrm{softmax}$ operation when drawing the probability distribution. 
To accelerate the translation, the widely used method is to trade off between the translation quality and the decoding speed by reducing the size of vocabulary \cite{mi2016coverage} or/and the number of parameters, which can not realize the full potential of NMT.

In this paper, we borrow ideas from phrase-based and syntax-based machine translation where 
cube pruning has been successfully applied to speed up the decoding \cite{chiang2007hierarchical,huang2007forest}.
Informally, cube pruning ``coarsens'' the search space by clustering similar states according to some equivalence relations.
To apply this idea to NMT, however, is much more involved.
Specifically, in the process of beam search, we cluster similar target hidden states to construct equivalence classes, the three dimensions of which are target words in the target vocabulary, part translations retained in the beam search and different combinations of similar target hidden states, respectively. The clustering operation can directly decrease the number of target hidden states in the following calculations, together with cube pruning, resulting in less RNN expansion operations to generate the next hidden state (related to the first reason) and less $\mathrm{softmax}$ operations over the target vocabulary (related to the second reason). 
The experiment results show that, when receiving the same or even better translation quality, our method can speed up the decoding speed by $3.3\times$ on GPUs and $3.5\times$ on CPUs.

\section{Background}
\label{sec2}

The proposed strategy can be adapted to optimize the beam search algorithm in the decoder of various NMT models. Without loss of generality, we take the attention-based NMT \cite{bahdanau2014neural} as an example to introduce our method. In this section, we first introduce the attention-based NMT model and then the cube pruning algorithm.

\subsection{The Attention-based NMT Model}

The attention-based NMT model follows the encoder-decoder framework with an extra attention module.
In the following parts, we will introduce each of the three components. Assume the source sequence and the observed translation are $\bm{\mathrm{x}}=\{x_1,\cdots,x_{|\bm{\mathrm{x}}|}\}$ and $\bm{\mathrm{y}}=\{y_1^{*},\cdots,y_{|\bm{\mathrm{y}}|}^{*}\}$.

{\bf Encoder}
The encoder uses a bidirectional GRU to obtain two sequences of hidden states. The final hidden state of each source word is got by concatenating the corresponding pair of hidden states in those sequences. Note that $e_{x_i}$ is employed to represent the embedding vector of the word $x_i$.
\begin{gather}
\overrightarrow{h}_i = \overrightarrow{\bm{\mathrm{GRU}}}\left(e_{x_i}, \overrightarrow{h}_{i-1}\right) \label{eq:left} \\
\overleftarrow{h}_i =  \overleftarrow{\bm{\mathrm{GRU}}}\left(e_{x_i}, \overleftarrow{h}_{i+1}\right) \label{eq:right} \\
h_i = \left[{\overrightarrow{h}_i};{\overleftarrow{h}_i}\right]   \label{eq:cat}
\end{gather}

{\bf Attention}
The attention module is designed to extract source information (called context vector) which is highly related to the generation of the next target word. At the $j$-th step, to get the context vector, the relevance between the target word $y_j^{*}$ and the $i$-th source word is firstly evaluated as
\begin{equation} \label{eq:att_query}
    r_{ij}=\bm{\mathrm{v}}_a^T \tanh\left(\bm{\mathrm{W}}_as_{j-1} + \bm{\mathrm{U}}_ah_i\right)
\end{equation}
Then, the relevance is normalized over the source sequence,
and all source hidden states are added weightedly to produce the context vector.
\begin{equation} \label{eq:att_weigth_sum}
    \alpha_{ij} = \frac{\exp \left( r_{ij} \right)}{\sum_{i'=1}^{|\bm{\mathrm{x}}|} \exp \left( r_{i'j} \right)};\ \ c_j = \sum\nolimits_{i=1}^{|\bm{\mathrm{x}}|}\alpha_{ij}h_i
\end{equation}

{\bf Decoder}
The decoder also employs a GRU to unroll the target information. The details are described in \citet{bahdanau2014neural}.
At the $j$-th decoding step, the target hidden state $s_j$ is given by
\begin{equation} \label{eq:s}
    s_j = f\left(e_{y_{j-1}^{*}}, s_{j-1}, c_j\right) \\
\end{equation}
The probability distribution $\mathcal{D}_{j}$ over all the words in the target vocabulary is predicted conditioned on the previous ground truth words, the context vector $c_j$ and the unrolled target information $s_{j}$.
\begin{gather}
t_j = g\left(e_{y_{j-1}^{*}}, c_j, s_j\right)  \label{eq:t} \\
o_j = \bm{\mathrm{W}}_o t_j \label{eq:o} \\
\mathcal{D}_{j} = \mathrm{softmax}\left(o_j\right)  \label{eq:softmax}
\end{gather}
where $g$ stands for a linear transformation, $\bm{\mathrm{W}}_o$ is used to map $t_j$ to $o_j$ so that each target word has one corresponding dimension in $o_j$.

\subsection{Cube Pruning}

\begin{figure*}[htp!]
    \centering
    \includegraphics[scale=0.60]{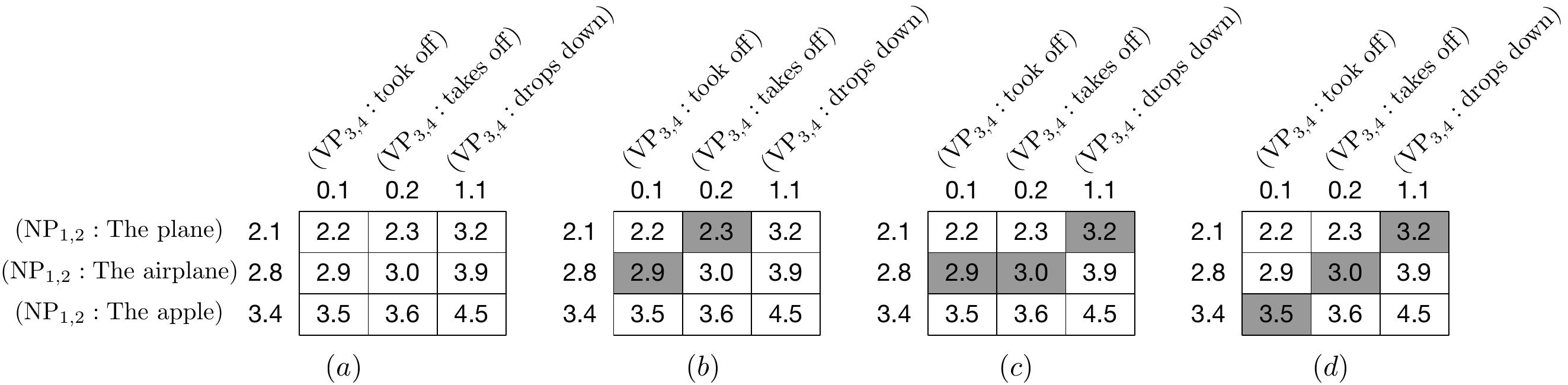}
    \caption{Cube pruning in SMT decoding. (a): the values in the grid denote the negative log-likelihood cost of the terminal combinations on both dimensions, and each dimension denotes a translation candidate in this example; (b)-(d): the process of popping the best candidate of top three items.}
    \label{fig:cpsmt}
\end{figure*}

\begin{table*}[!htbp]
\centering
\renewcommand\arraystretch{1.2}
\begin{tabular}{c||c|c||c|c}
    					& \multicolumn{2}{c||}{\bf GPU}	&	\multicolumn{2}{c}{\bf CPU}\\
    \hline
	{\bf Calculation Units}     & {\bf Time(s)}  & {\bf Percentage} & {\bf Time(s)}    & {\bf Percentage}	\\
	\hline
	Eq. (\ref{eq:s}): $s_j = f(e_{y_{j-1}^{*}}, s_{j-1}, c_j)$ 		 & 551.07    & 75.73\%  & 1370.92     & 19.42\% \\
	Eq. (\ref{eq:t}): $t_j = g(e_{y_{j-1}^{*}}, c_j, s_j)$ 		 	 & 88.25     & 12.13\%  & 277.76      & 3.93\% 	\\
	Eq. (\ref{eq:o}): $o_j = \bm{\mathrm{W}}_o t_j$ 		 & 25.33       & 3.48\%  	& 2342.53     & 33.18\%	\\
	Eq. (\ref{eq:softmax}): $\mathcal{D}_{j} = \mathrm{softmax}(o_j)$      & 63.00     & 8.66\%  	& 3069.25     & 43.47\%\\
\end{tabular}
\caption{Time cost statistics for decoding the whole MT03 testset on GPUs and CPUs with beam size $10$.}
\label{table:time_statistics}
\end{table*}

The cube pruning algorithm, proposed by \citet{chiang2007hierarchical} based on the $k$-best parsing algorithm of \citet{huang+chiang:2005}, is actually an accelerated extension based on the naive beam search algorithm. Beam search, a heuristic dynamic programming searching algorithm, explores a graph by expanding the most promising nodes in a limited set and searches approximate optimal results from candidates. For the sequence-to-sequence learning task, given a pre-trained model, the beam search algorithm finds a sequence that approximately maximizes the conditional probability \cite{graves2012sequence,boulanger2013audio}.
Both \citet{sutskever2014sequence} and \citet{bahdanau2014neural} employed the beam search algorithm into the NMT decoding to produce translations with relatively larger conditional probability with respect to the optimized model parameters. Remarkably, \citet{huang2007forest} successfully applied the cube pruning algorithm to the decoding of SMT. They found that the beam search algorithm in SMT can be extended, and they utilized the cube pruning and some variants to optimize the search process in the decoding phase of phrase-based \cite{och2004alignment} and syntax-based \cite{chiang2005hierarchical,galley2006scalable} systems, which decreased a mass of translation candidates and achieved a significant speed improvement by reducing the size of complicated search space, thereby making it possible to actualize the thought of improving the translation performance through increasing the beam size.

In the traditional SMT decoding, the cube pruning algorithm aims to prune a great number of partial translation hypotheses without computing and storing them. For each decoding step, those hypotheses with the same translation rule are grouped together, then the cube pruning algorithm is conducted over the hypotheses. We illustrate the detailed process in Figure \ref{fig:cpsmt}.

\section{NMT Decoder with Cube Pruning}
\label{sec3}

\subsection{Definitions}
We define the related storage unit tuple of the $i$-th candidate word in the $j$-th beam as $n_j^{i} = (c_j^{i}, s_j^{i}, y_j^{i}, bp_j^{i})$, where $c_j^{i}$ is the negative log-likelihood (NLL) accumulation in the $j$-th beam, $s_j^{i}$ is the decoder hidden state in the $j$-th beam, $y_j^{i}$ is the index of the $j$-th target word in large vocabulary and $bp_j^{i}$ is the backtracking pointer for the $j$-th decoding step. 
Note that, for each source sentence, we begin with calculating its encoded representation and the first hidden state $s_0^{0}$ in decoder, then searching from the initial tuple $(0.0, s_0^{0}, 0, 0)$ existing in the first beam\footnote{\noindent The initial target word index $y_0^{0}$ equals to $0$, which actually corresponds to the Beginning Of Sentence (BOS) token in target vocabulary.}.

It is a fact that Equation (\ref{eq:softmax}) produces the probability distribution of the predicted target words over the target vocabulary $V$. \citet{cho2014learning} indicated that whenever a target word is generated, the $\mathrm{softmax}$ function over $V$ computes probabilities for all words in $V$, so the calculation is expensive when the target vocabulary is large. As such, \citet{bahdanau2014neural} (and many others) only used the top-$30k$ frequent words as target vocabulary, and replaced others with UNK. However, the final normalization operation still brought high computation complexity for forward calculations.

\subsection{Time Cost in Decoding}

We conducted an experiment to explore how long each calculation unit in the decoder would take. We decoded the MT03 test dataset by using naive beam search with beam size of $10$ and recorded the time consumed in the computation of Equation (\ref{eq:s}), (\ref{eq:t}), (\ref{eq:o}) and (\ref{eq:softmax}), respectively. The statistical results in Table \ref{table:time_statistics} show that the recurrent calculation unit consumed the most time on GPUs, while the $\mathrm{softmax}$ computation also took lots of time. On CPUs, the most expensive computational time cost was caused by the $\mathrm{softmax}$ operation over the entire target vocabulary\footnote{Note that, identical to \citet{bahdanau2014neural}, we only used $30k$ as the vocabulary size.}. In order to avoid the time-consuming normalization operation in testing, we introduced \textit{self-normalization} (denoted as {\bf SN}) into the training.

\subsection{Self-normalization}

\textit{Self-normalization}~\cite{devlin2014selfnorm} was designed to make the model scores which are produced by the output layer be approximated by the probability distribution over the target vocabulary without normalization operation. According to Equation (\ref{eq:softmax}), for an observed target sentence $\bm{\mathrm{y}}=\{y_1^{*},\cdots,y_{|\bm{\mathrm{y}}|}^{*}\}$, the Cross-Entropy (CE) loss could be written as
\begin{equation} \label{eq:sn:0}
\begin{aligned}
\mathrm{L}_\theta &= -\sum_{j=1}^{|\bm{\mathrm{y}}|}\log \mathcal{D}_{j}[y_j^{*}]\\
		 		  &= -\sum_{j=1}^{|\bm{\mathrm{y}}|}\log\frac{\exp\left(o_j[y_j^{*}]\right)}{\sum\nolimits_{y^{\prime}\in V}\exp\left(o_j[y^{\prime}]\right)}\\
         		  &= \sum_{j=1}^{|\bm{\mathrm{y}}|}\log\sum_{y^{\prime}\in V}\exp\left(o_j[y^{\prime}]\right)-o_j[y_j^{*}]
\end{aligned}
\end{equation}
where $o_j$ is the model score generated by Equation (\ref{eq:o}) at the $j$-th step, we marked the $\mathrm{softmax}$ normalizer $\sum\nolimits_{y^{\prime}\in V}\exp\left(o_j[y^{\prime}]\right)$ as $Z$.

Following the work of \citet{devlin2014selfnorm}, we modified the CE loss function into 
\begin{equation} \label{eq:sn:1}
\begin{aligned}
\mathrm{L}_\theta &= -\sum_{j=1}^{|\bm{\mathrm{y}}|}\left(\log \mathcal{D}_{j}[y_j^{*}] - \alpha(\log Z - 0)^2\right)\\
				  &= -\sum_{j=1}^{|\bm{\mathrm{y}}|}\left(\log \mathcal{D}_{j}[y_j^{*}] - \alpha\log^2Z\right)
\end{aligned}
\end{equation}

The objective function, shown in Equation (\ref{eq:sn:1}), is optimized to make sure $\log Z$ is approximated to $0$, equally, make $Z$ close to $1$ once it converges. We chose the value of $\alpha$ empirically. Because the $\mathrm{softmax}$ normalizer $Z$ is converged to $1$ in inference, we just need to ignore $Z$ and predict the target word distribution at the $j$-th step only with $o_j$:

\begin{equation} \label{eq:sn:2}
\begin{aligned}
\mathcal{D}_{j} = o_j
\end{aligned}
\end{equation}

\subsection{Cube Pruning}

\begin{figure*}[!t]
    \centering
    \includegraphics[scale=0.70]{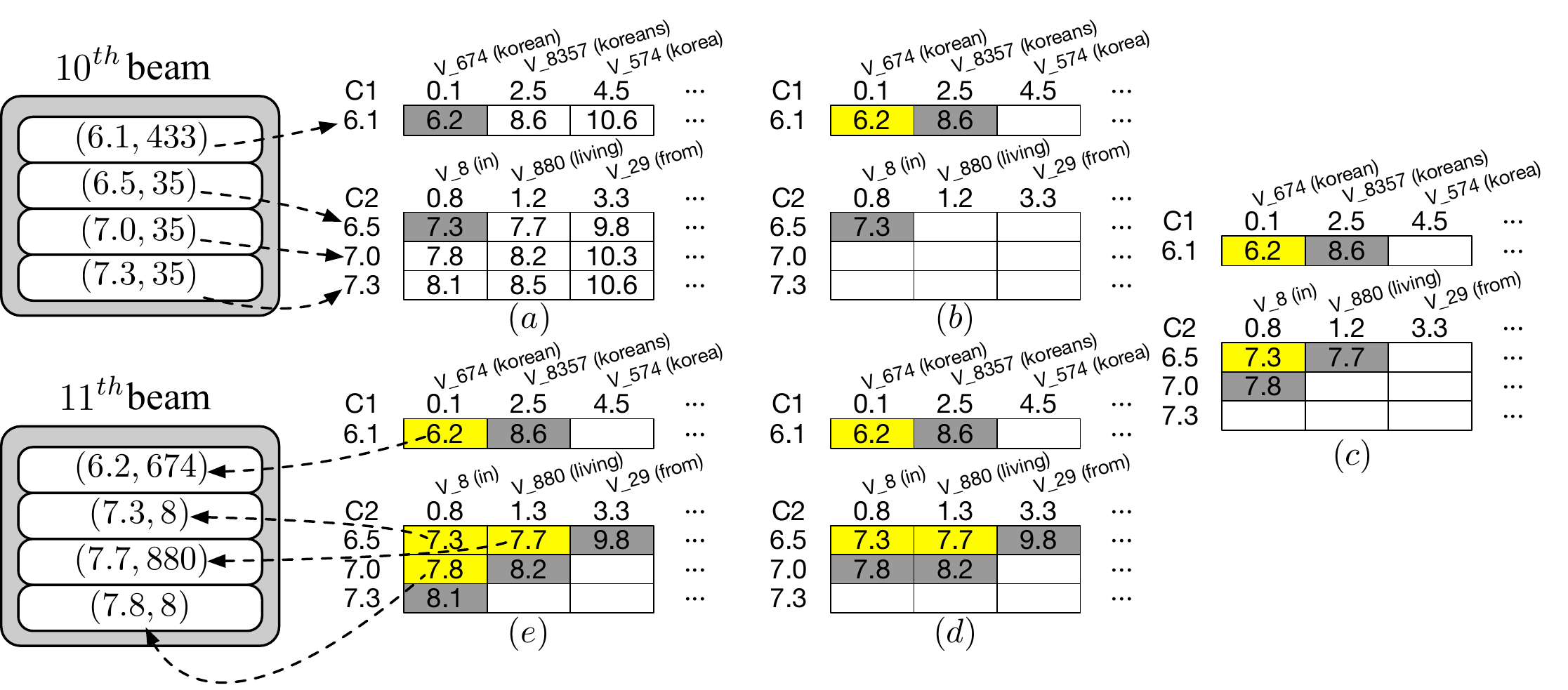}
    \caption{Cube pruning diagram in beam search process during NMT decoding. We only depict the accumulated NLL and the word-level candidate for each item in the beam (in the bracket). Assume the beam size is $4$, we initialize a heap for the current step, elements in the $10^{th}$ beam are merged into two sub-cubes $\mathrm{C}1$ and $\mathrm{C}2$ according to the previous target words; (a) the two elements located in the upper-left corner of the two sub-cubes are pushed into the heap; (b) minimal element $(6.2, 674)$ is popped out, meanwhile, its neighbor $(8.6, 8357)$ is pushed into the heap; (c) minimal element $(7.3, 8)$ is popped out, its right-neighbor $(7.7, 880)$ and lower-neighbor $(7.8, 8)$ are pushed into the heap; (d) minimal element $(7.7, 880)$ is popped out, its right-neighbor $(9.8, 29)$ and down-neighbor $(8.2, 880)$ are pushed into the heap; (e) minimal element $(7.8, 8)$ is popped out, then its down-neighbor $(8.1, 8)$ is pushed into the heap. $4$ elements have been popped out, we use them to construct the $11^{th}$ beam. Yellow boxes indicate the $4$-best word-level candidates to be pushed into the $11^{th}$ beam.}
    \label{fig:cubepruning}
\end{figure*}

Table \ref{table:time_statistics} clearly shows that the equations in the NMT forward calculation take lots of time. Here, according to the idea behind the cube pruning algorithm, we tried to reduce the time of time-consuming calculations, e.g., Equation (\ref{eq:s}), and further decrease the search space by introducing the cube pruning algorithm.

\subsubsection{Integrating into NMT Decoder} 	\label{icpibs}
Extended from the naive beam search in the NMT decoder, cube pruning, treated as a pruning algorithm, attempts to reduce the search space and computation complexity by merging some similar items in a beam to accelerate the naive beam search, keeping the $1$-best searching result almost unchanged or even better by increasing the beam size. Thus, it is a fast and effective algorithm to generate candidates.

Assume that $\mathrm{T}$ restores the set of the finished translations. For each step in naive beam search process, $\mathrm{beamsize}$$-$$|\mathrm{T}|$ times forward calculations are required to acquire $\mathrm{beamsize}$$-$$|\mathrm{T}|$ probability distributions corresponding to each item in the previous beam \cite{bahdanau2014neural}. while for each step in cube pruning, in terms of some constraints, we merge all similar items in the previous beam into one {\em equivalence class} (called a sub-cube). The constraint we used here is that items being merged in the previous beam should have the same target words. Then, for the sub-cube, only one forward calculation is required to obtain the approximate predictions by using the loose hidden state. Elements in the sub-cube are sorted by previous accumulated NLL along the columns (the first dimension of beam size) and by the approximate predictions along the rows (the second dimension of vocabulary size). After merging, one beam may contain several sub-cubes (the third dimension), we start to search from item in the upper left corner of each sub-cube, which is the best one in the sub-cube, and continue to spread out until enough candidates are found. Once a item is selected, the exact hidden state will be used to calculate its exact NLL.

Through all above steps, the frequency of forward computations decreases. We give an example to dive into the details in Figure \ref{fig:cubepruning}.

Assume that the beam size is $4$. Given the $10^{th}$ beam, we generate the $11^{th}$ beam. Different from the naive beam search, we first group items in the previous beam into two sub-cubes $\mathrm{C}1$ and $\mathrm{C}2$ in term of the target word $y_{j-1}$. As shown in part $(a)$ of Figure \ref{fig:cubepruning}, $(6.1, 433)$ constructs the sub-cube $\mathrm{C}1$; $(6.5, 35)$, $(7.0, 35)$ and $(7.3, 35)$ are put together to compose another sub-cube $\mathrm{C}2$. Items in part $(a)$ are ranked in ascending order along both row and column dimension according to the accumulated NLL. For each sub-cube, we use the first state vector in each sub-cube as the approximate one to produce the next probability distribution and the next state.
At beginning, each upper-left corner element in each sub-cube is pushed into a minimum heap, after popping minimum element from the heap, we calculate and restore the exact NLL of the element, then push the right and lower ones alongside the minimum element into heap. At this rate, the searching continues just like the ``diffusion'' in the sub-cube until $4$ elements are popped, which are ranked in terms of their exact NLLs to construct the $11^{th}$ beam. Note that once an element is popped, we calculate its exact NLL. From the step (e) in Figure \ref{fig:cubepruning}, we can see that $4$ elements have been popped from $\mathrm{C}1$ and $\mathrm{C}2$, and then ranked in terms of their exact NLLs to build the $11^{th}$ beam.

We refer above algorithm as the naive cube pruning algorithm (called {\bf NCP})

\subsubsection{Accelerated Cube Pruning}

In each step of the cube pruning algorithm, after merging the items in the previous beam, some similar candidates are grouped together into one or more sub-cube(s). We also try to predict the approximate distribution for each sub-cube only according to the top-$1$ state vector (the first row in the sub-cube in Figure \ref{fig:cubepruning}), and select next candidates after ranking. The predicted probability distribution will be very similar to that of the naive beam search. Besides, Each sub-cube only requires one forward calculation. Thus, it could save more search space and further reduce the computation complexity for the decoder. Unlike the naive cube pruning algorithm, accelerated cube pruning pops each item, then still use the approximate NLL instead of the exact one. We denote this kind of accelerated cube pruning algorithm as {\bf ACP}.

\section{Experiments}
\label{sec4}
We verified the effectiveness of proposed cube pruning algorithm on the Chinese-to-English (Zh-En) translation task.
\subsection{Data Preparation}
The Chinese-English training dataset consists of $1.25$M sentence pairs\footnote{These sentence pairs are mainly extracted from LDC2002E18, LDC2003E07, LDC2003E14, Hansards portion of LDC2004T07, LDC2004T08 and LDC2005T06}. We used the NIST 2002 (MT02) dataset as the validation set with $878$ sentences, and the NIST 2003 (MT03) dataset as the test dataset, which contains $919$ sentences.

The lengths of the sentences on both sides were limited up to $50$ tokens, then actually $1.11$M sentence pairs were left with $25.0$M Chinese words and $27.0$M English words. We extracted $30k$ most frequent words as the source and target vocabularies for both sides.

In all the experiments, case-insensitive $4$-gram BLEU \cite{papineni2002bleu} was employed for the automatic evaluation, we used the script \emph{mteval-v11b.pl}\footnote{\scriptsize \url{https://github.com/moses-smt/mosesdecoder/blob/master/scripts/generic/mteval-v11b.pl}} to calculate the BLEU score.

\subsection{System}
The system is an improved version of attention-based NMT system named RNNsearch~\cite{bahdanau2014neural} where the decoder employs a conditional GRU layer with attention, consisting of two GRUs and an attention module for each step\footnote{\scriptsize \url{https://github.com/nyu-dl/dl4mt-tutorial/blob/master/docs/cgru.pdf}}. Specifically, Equation (\ref{eq:s}) is replaced with the following two equations:
\begin{gather} \label{imp_dec}
    \tilde{s}_j = \bm{\mathrm{GRU}}_1(e_{y_{j-1}^{*}}, s_{j-1}) \\
    s_j = \bm{\mathrm{GRU}}_2(c_j, \tilde{s}_j)
\end{gather}
Besides, for the calculation of relevance in Equation (\ref{eq:att_query}), $s_{j-1}$ is replaced with $\tilde{s}_{j-1}$. The other components of the system keep the same as RNNsearch.
Also, we re-implemented the beam search algorithm as the naive decoding method, and naive searching on the GPU and CPU server were conducted as two baselines.

\subsection{Training Details}
Specially, we employed a little different settings from \citet{bahdanau2014neural}: Word embedding sizes on both sides were set to $512$, all hidden sizes in the GRUs of both encoder and decoder were also set to $512$. All parameter matrices, including bias matrices, were initialized with the uniform distribution over $\left[-0.1,0.1\right]$. Parameters were updated by using mini-batch Stochastic Gradient Descent (SGD) with batch size of $80$ and the learning rate was adjusted by AdaDelta \cite{zeiler2012adadelta} with decay constant $\rho$=$0.95$ and denominator constant $\epsilon$=$1e\textnormal{-}6$. The gradients of all variables whose $\mathrm{L}2$-norm are larger than a pre-defined threshold $1.0$ were normalized to the threshold to avoid gradient explosion \cite{pascanu2013difficulty}. Dropout was applied to the output layer with dropout rate of $0.5$. We exploited length normalization \cite{cho2014properties} strategy on candidate translations in beam search decoding.

The model whose BLEU score was the highest on the validation set was used to do testing. Maximal epoch number was set to $20$. Training was conducted on a single Tesla K80 GPU, it took about $2$ days to train a single NMT model on the Zh-En training data. For \textit{self-normalization}, we empirically set $\alpha$ as $0.5$ in Equation (\ref{eq:sn:1})\footnote{Following \citet{devlin2014selfnorm}, we had tried $0.01$, $0.1$, $0.5$ $1.0$ and $10.0$ for the value of $\alpha$, we found that $0.5$ produced the best result.}.

\subsection{Search Strategies}
We conducted experiments to decode the MT03 test dataset on the GPU and CPU server respectively, then compared search quality and efficiency among following six search strategies under different beam sizes.

{\bf NBS-SN:} Naive Beam Search without {\bf SN}

{\bf NBS+SN:} Naive Beam Search with {\bf SN}

{\bf NCP-SN:} Cube Pruning without {\bf SN}

{\bf NCP+SN:} Cube Pruning with {\bf SN}

{\bf ACP-SN:} Accelerated Cube Pruning without {\bf SN}

{\bf ACP+SN:} Accelerated Cube Pruning with {\bf SN}

\begin{figure}[!hbt]
    \centering
    \includegraphics[scale=0.6]{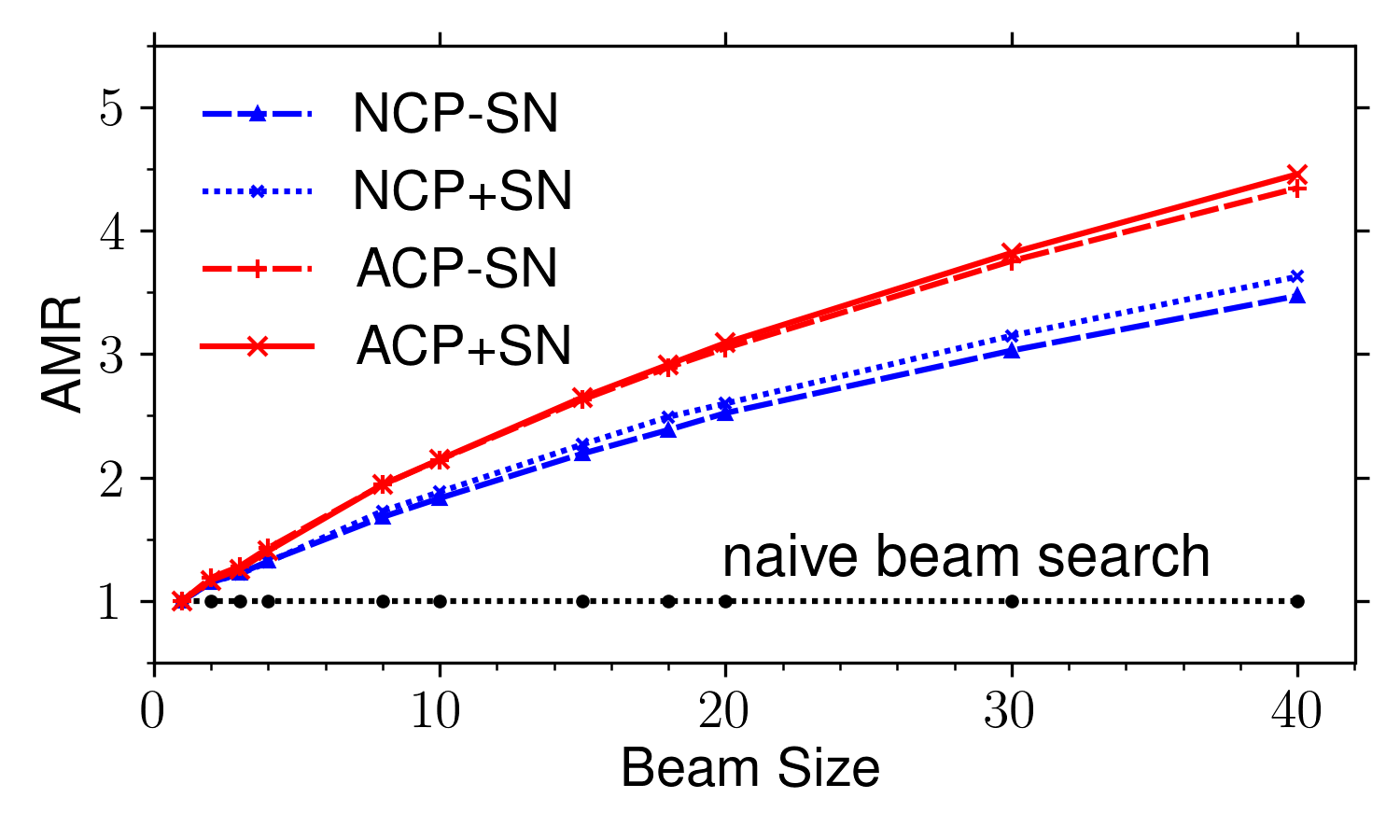}
    \caption{AMR comparison on the MT03 test dataset. Decoding the MT03 test dataset on a single GeForce GTX TITAN X GPU server under the different searching settings. y-axis represents the AMR on the test dataset in the whole searching process and x-axis indicates beam size. Unsurprisingly, we got exactly the same results on the CPU server, not shown here.}
    \label{fig:avg_mrate_vs_beamsize_nist03}
\end{figure}

\subsection{Comparison of Average Merging Rate} 	\label{sec4.2}

We first give the definition of the Average Merging Rate (denoted as AMR). Given a test dataset, we counted the total word-level candidates (noted as $\mathrm{N}_w$) and the total sub-cubes (noted as $\mathrm{N}_c$) during the whole decoding process, then the AMR can be simply computed as
\begin{equation} \label{eq:amr}
    \overline{m}=\mathrm{N}_w/\mathrm{N}_c
\end{equation}

The MT03 test dataset was utilized to compare the trends of the AMR values under all six methods. We used the pre-trained model to translate the test dataset on a single GeForce GTX TITAN X GPU server. Beam size varies from $1$ to $40$, values are included in the set $\{ 1, 2, 3, 4, 8, 10, 15, 18, 20, 30, 40\}$. For each beam size, six different searching settings were applied to translate the test dataset respectively.
The curves of the AMRs during the decoding on the MT03 test dataset under the proposed methods are shown in Figure \ref{fig:avg_mrate_vs_beamsize_nist03}. Note that the AMR values of {\bf NBS} are always $1$ whether there is {\bf SN} or not.

Comparing the curves in the Figure \ref{fig:avg_mrate_vs_beamsize_nist03}, we could observe that the naive beam search does not conduct any merging operation in the whole searching process, while the average merging rate in the cube pruning almost grows as the beam size increases. Comparing the red curves to the blue ones, we can conclude that, in any case of beam size, the AMR of the accelerated cube pruning surpasses the basic cube pruning by a large margin. Besides, self-normalization could produces the higher average merging rate comparing to the counterpart without self-normalization.

\begin{figure*}[t!]
\centering
\begin{subfigure}{.48\textwidth}
  \centering
  \includegraphics[width=1.0\linewidth]{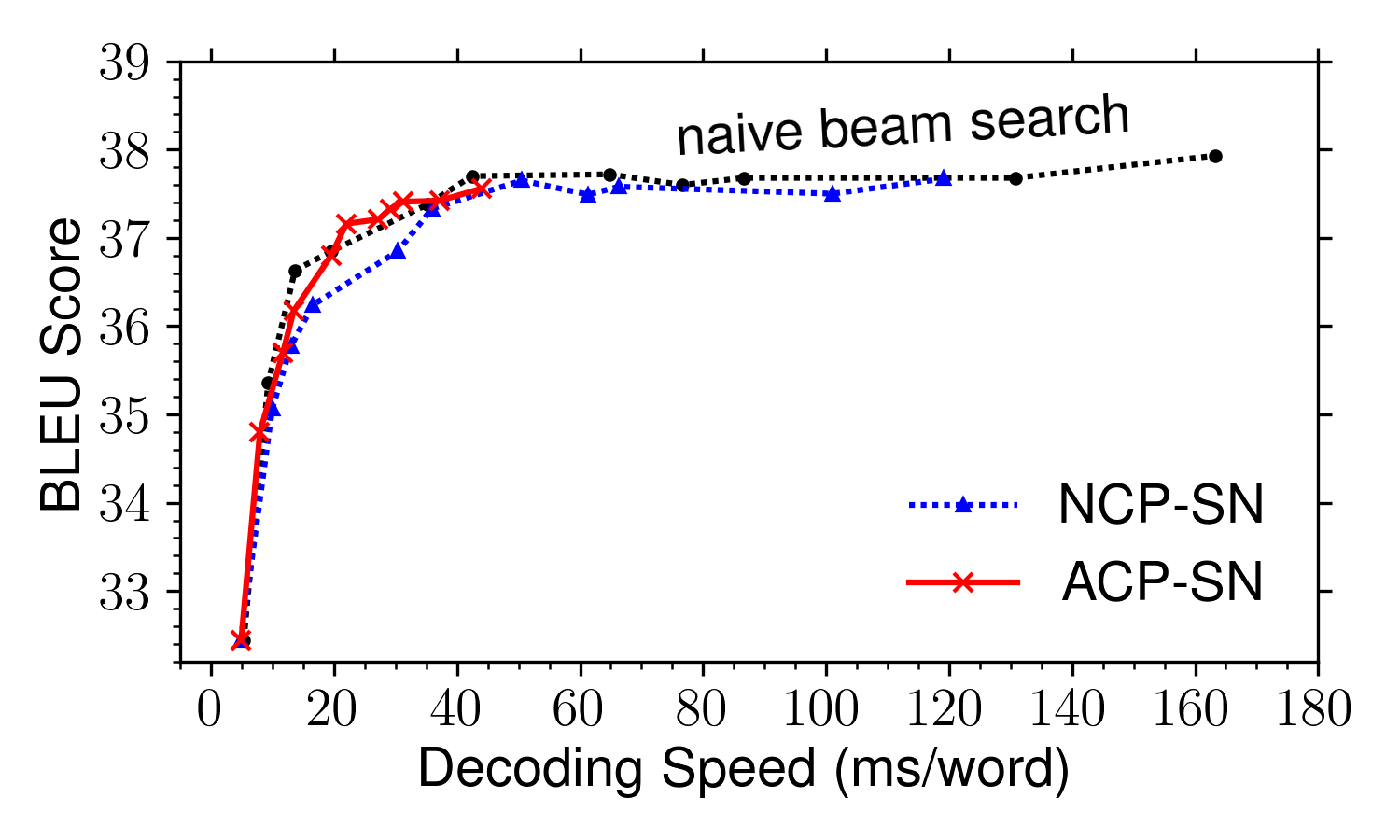}
  \caption{BLEU vs.~decoding speed, without self-normalization}
  \label{fig:gpu-nist03-osn}
\end{subfigure}%
\quad
\begin{subfigure}{.48\textwidth}
  \centering
  \includegraphics[width=1.0\linewidth]{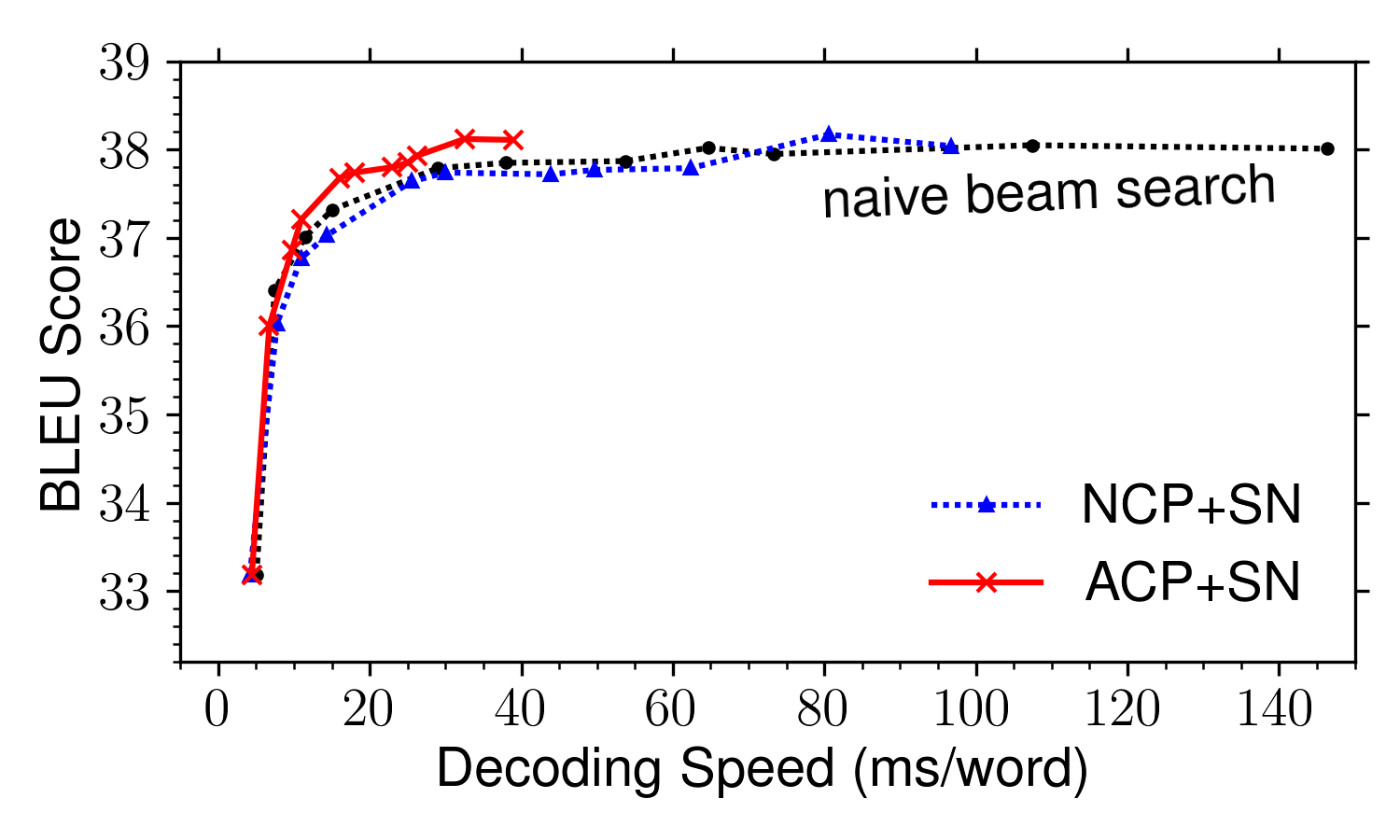}
  \caption{BLEU vs.~decoding speed, with self-normalization}
  \label{fig:gpu-nist03-sn}
\end{subfigure}
\caption{Comparison among the decoding results of the MT03 test dataset on the single GeForce GTX TITAN X GPU server under the three different searching settings. y-axis represents the BLEU score of translations, x-axis indicates that how long it will take for translating one word on average.}
\label{fig:compare_gpu_nist03}
\end{figure*}

\subsection{Comparison on the GPU Server}
Intuitively, as the value of the AMR increases, the search space will be reduced and computation efficiency improves. 
We compare the two proposed searching strategies and the naive beam search in two conditions (with self-normalization and without self-normalization). 
Figure \ref{fig:compare_gpu_nist03} demonstrates the results of comparison between the  proposed searching methods and the naive beam search baseline in terms of search quality and search efficiency under different beam sizes.

By fixing the beam size and the dataset, we compared the changing trend of BLEU scores for the three distinct searching strategies under two conditions. Without self-normalization, Figure \ref{fig:gpu-nist03-osn} shows the significant improvement of the search speed, however the BLEU score drops about $0.5$ points. We then equipped the search algorithm with self-normalization. Figure \ref{fig:gpu-nist03-sn} shows that the accelerated cube pruning search algorithm only spend about one-third of the time of the naive beam search to achieve the best BLEU score with beam size $30$. Concretely, when the beam size is set to be $30$, {\bf ACP+SN} is $3.3$ times faster than the baseline on the MT03 test dataset, and both performances are almost the same.

\begin{figure*}[t!]
\centering
\begin{subfigure}{.48\textwidth}
  \centering
  \includegraphics[width=1.0\linewidth]{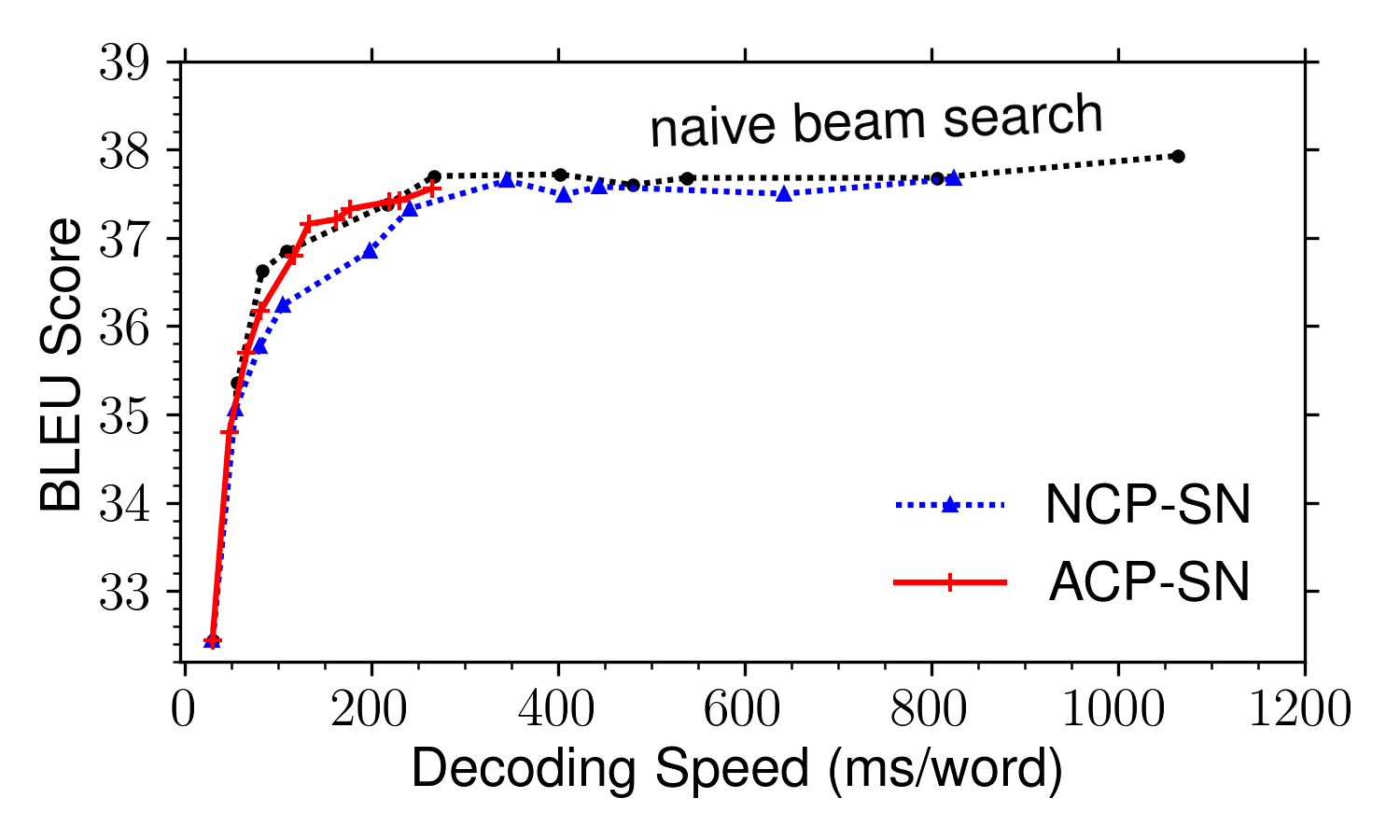}
  \caption{BLEU vs.~decoding speed, without self-normalization}
  \label{fig:cpu-nist03-osn}
\end{subfigure}%
\quad
\begin{subfigure}{.48\textwidth}
  \centering
  \includegraphics[width=1.0\linewidth]{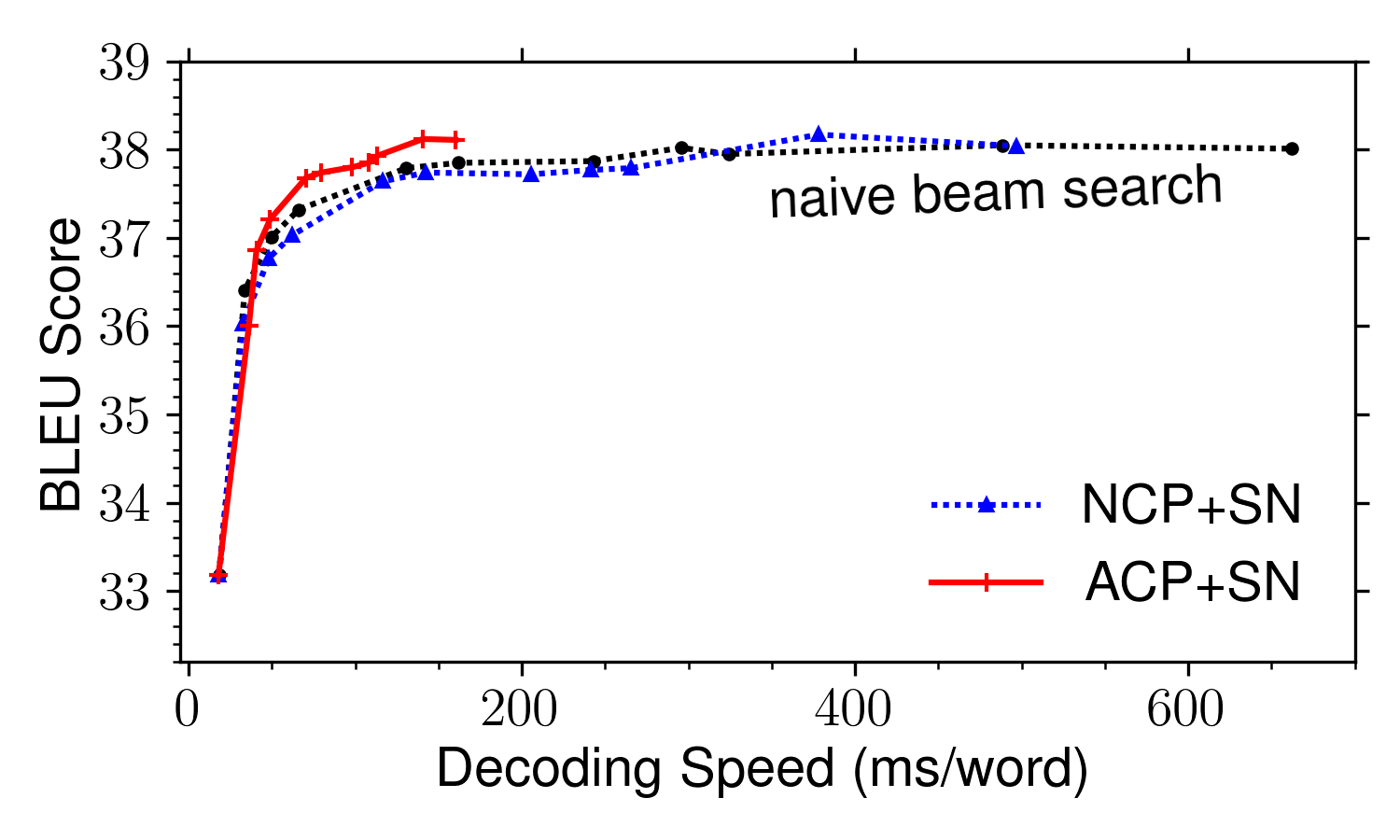}
  \caption{BLEU vs.~decoding speed, with self-normalization}
  \label{fig:cpu-nist03-sn}
\end{subfigure}
\caption{Comparison among the decoding results of the MT03 test dataset on the single AMD Opteron(tm) Processor under the three different searching settings. y-axis represents the BLEU score of translations, x-axis indicates that how long it will take for translating one word on average.}
\label{fig:compare_cpu_nist03}
\end{figure*}

\subsection{Comparison on the CPU Server}

Similar to the experiments conducted on GPUs, we also translated the whole MT03 test dataset on the CPU server by using all six search strategies under different beam sizes.
The trends of the BLEU scores over those strategies are shown in Figure \ref{fig:compare_cpu_nist03}.

The proposed search methods gain the similar superiority on CPUs to that on GPUs, and the decoding speed is obviously slower than that on GPUs. From the Figure \ref{fig:cpu-nist03-osn}, we can also clearly see that, compared with the {\bf NBS-SN}, {\bf NCP-SN} only speeds up the decoder a little, {\bf ACP-SN} produces much more acceleration. However, when we did not introduce self-normalization, the proposed search methods will also result in a loss of about $0.5$ BLEU score. The self-normalization made the {\bf ACP} strategy faster than the baseline by about $3.5\times$, in which condition the {\bf NBS+SN} got the best BLEU score $38.05$ with beam size $30$ while the {\bf ACP+SN} achieved the highest score $38.12$ with beam size $30$. The results could be observed in Figure \ref{fig:cpu-nist03-sn}. Because our method is on the algorithmic level and platform-independent, it is reasonable that the proposed method can not only perform well on GPUs, but also accelerate the decoding significantly on CPUs. Thus, the accelerated cube pruning with self-normalization could improve the search quality and efficiency stably.

\subsection{Decoding Time}

\begin{figure*}[t!]
\centering
\begin{subfigure}{.48\textwidth}
  \centering
  \includegraphics[width=1.0\linewidth]{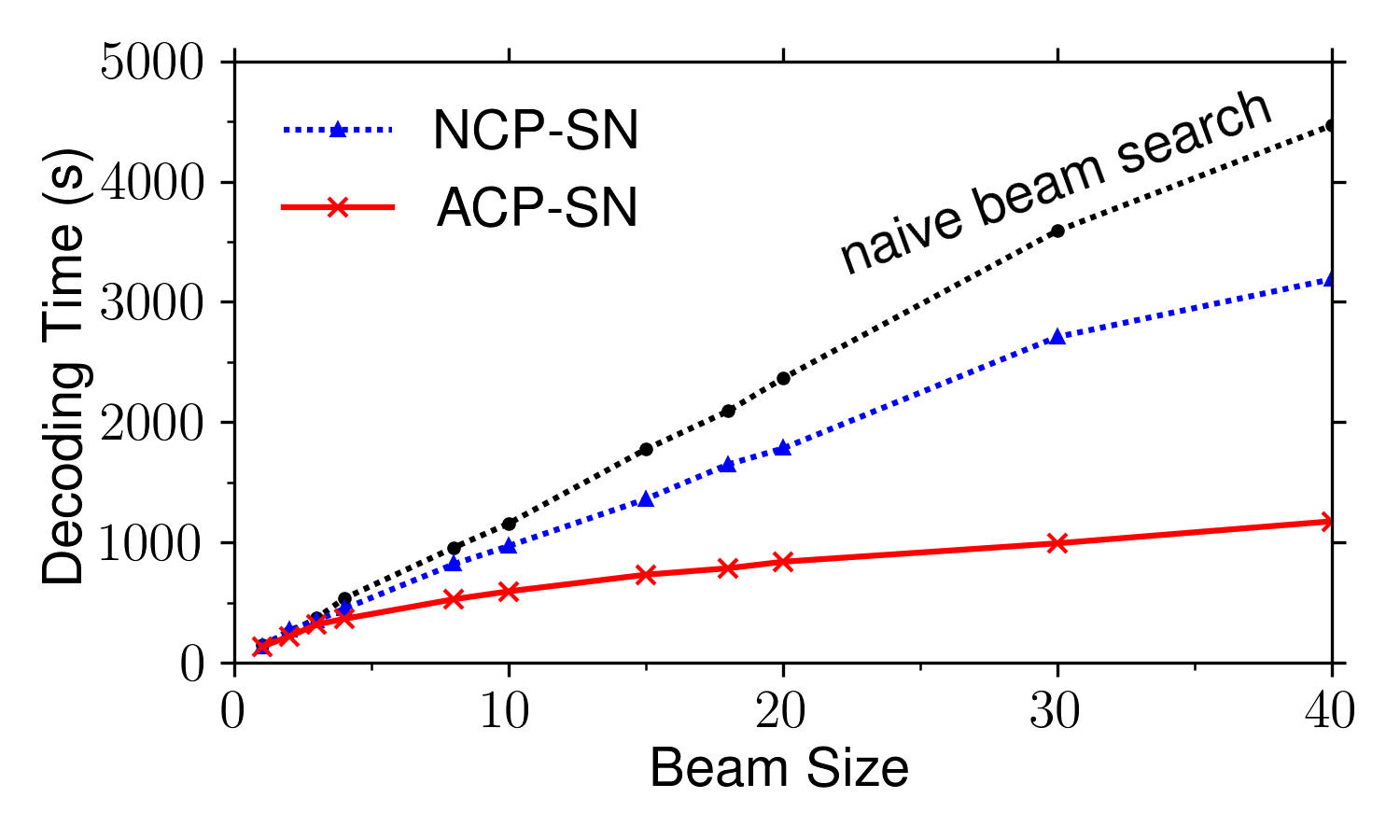}
  \caption{Time spent on translating MT03 test dataset for different beam sizes without self-normalization}
  \label{fig:gpu-nist03-osn-time}
\end{subfigure}%
\quad
\begin{subfigure}{.48\textwidth}
  \centering
  \includegraphics[width=1.0\linewidth]{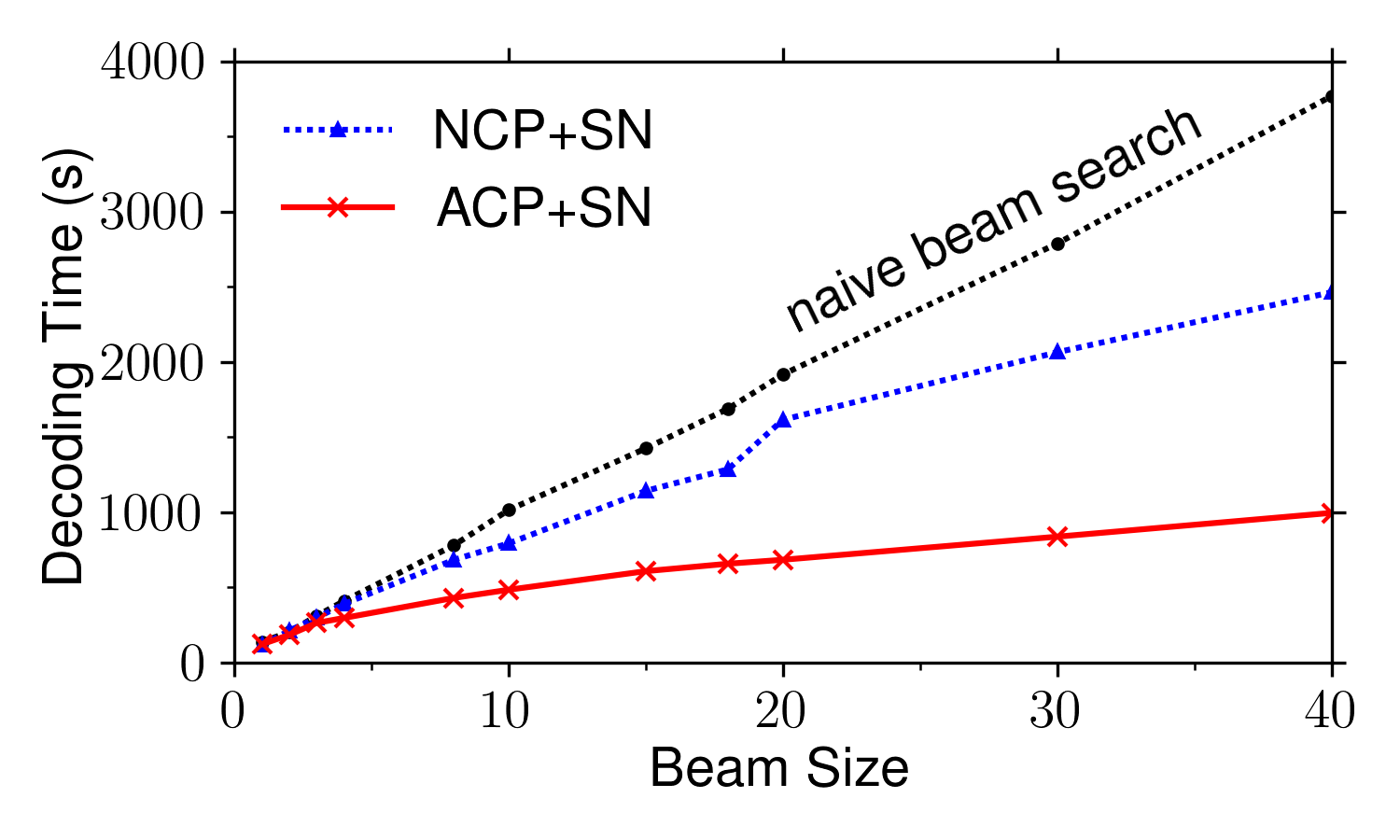}
  \caption{Time spent on translating MT03 test dataset for different beam sizes with self-normalization}
  \label{fig:gpu-nist03-sn-time}
\end{subfigure}
\caption{Comparison among the decoding results of the MT03 test dataset on the single GeForce GTX TITAN X GPU server under the three different searching settings. y-axis represents the BLEU score of translations, x-axis indicates that how long it will take for translating one word on average.}
\label{fig:compare_gpu_decoing_time}
\end{figure*}

\begin{figure*}[t!]
\centering
\begin{subfigure}{.48\textwidth}
  \centering
  \includegraphics[width=1.0\linewidth]{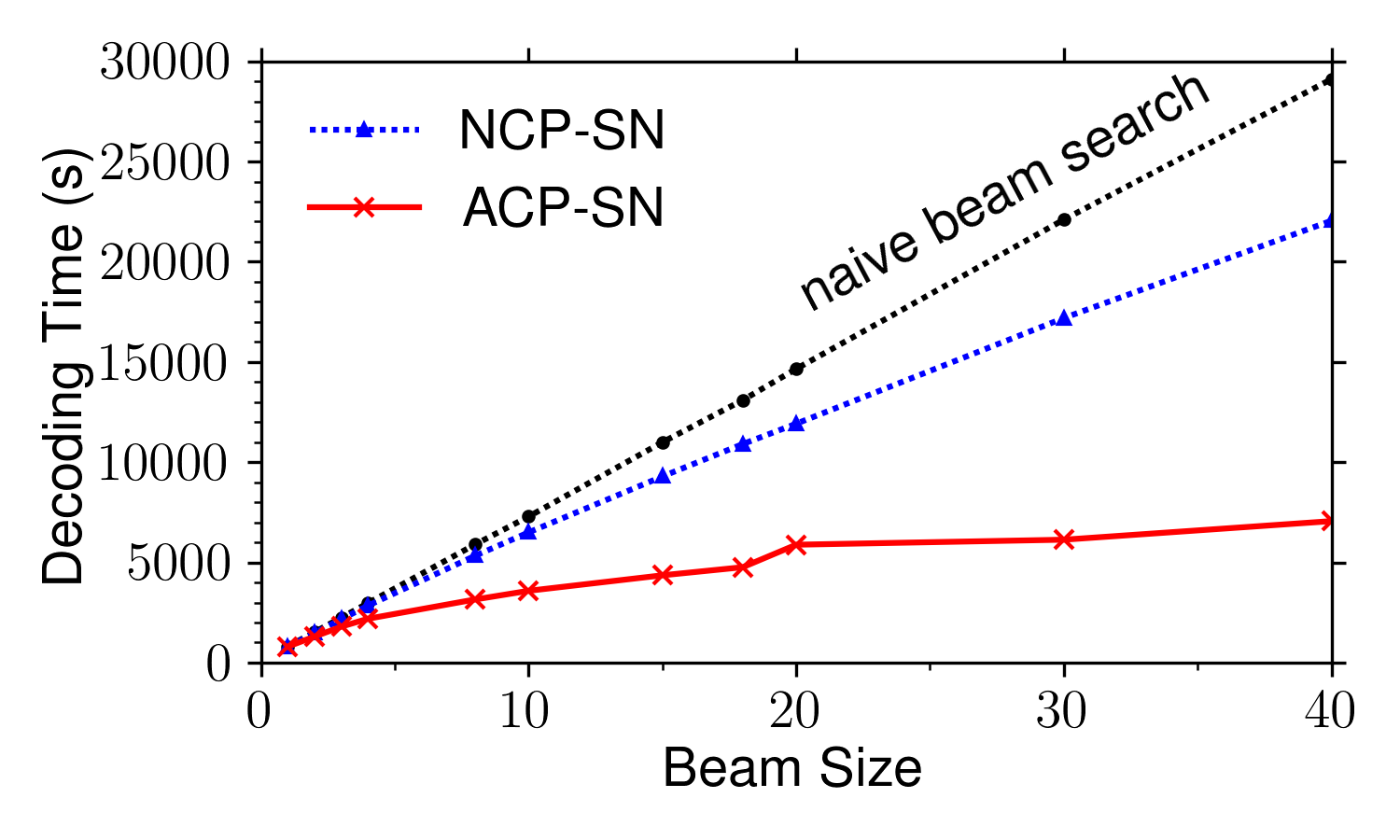}
  \caption{Time spent on translating MT03 test dataset for different beam sizes without self-normalization}
  \label{fig:cpu-nist03-osn-time}
\end{subfigure}%
\quad
\begin{subfigure}{.48\textwidth}
  \centering
  \includegraphics[width=1.0\linewidth]{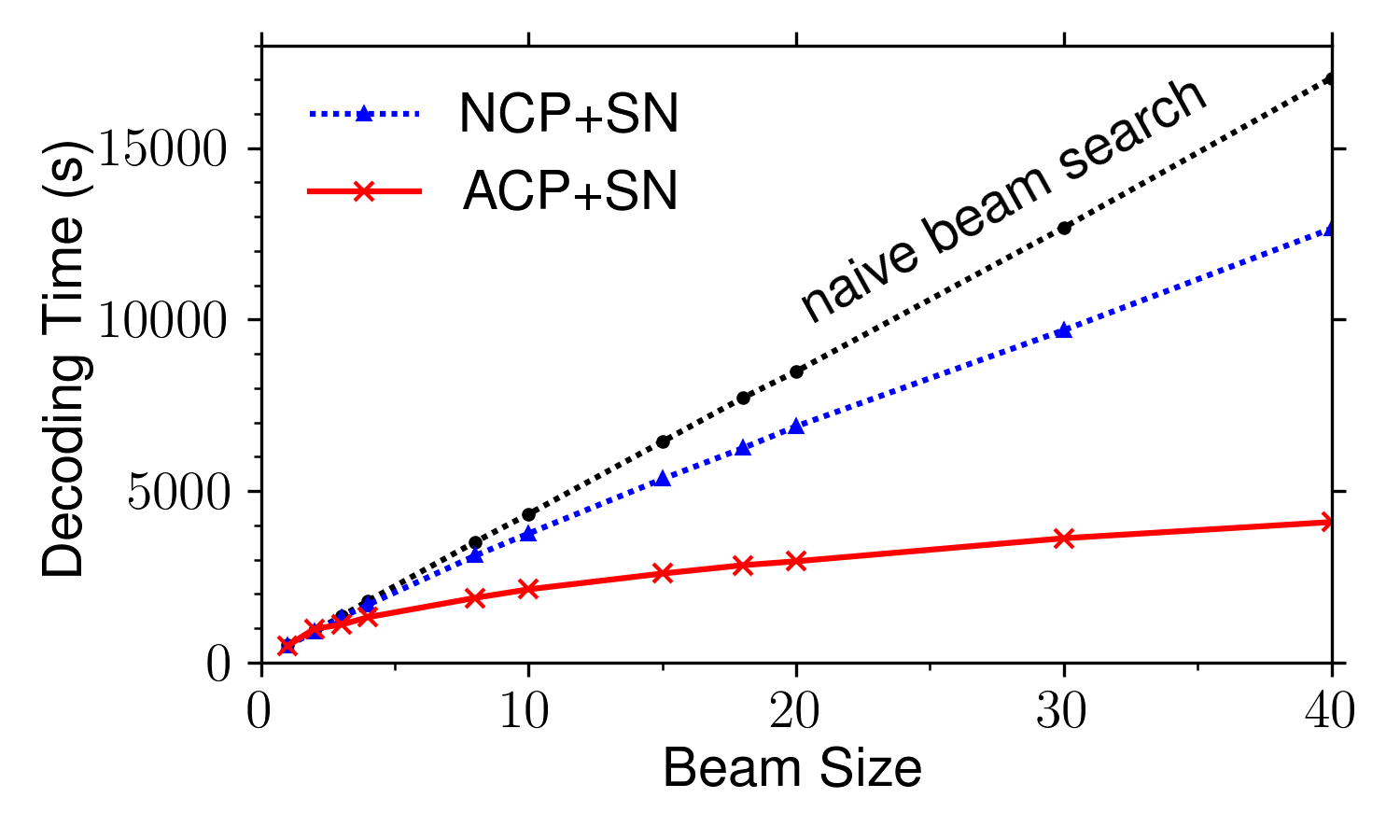}
  \caption{Time spent on translating MT03 test dataset for different beam sizes with self-normalization}
  \label{fig:cpu-nist03-sn-time}
\end{subfigure}
\caption{Comparison among the decoding results of the MT03 test dataset on the single AMD Opteron(tm) Processor under the three different searching settings. y-axis represents the BLEU score of translations, x-axis indicates that how long it will take for translating one word on average.}
\label{fig:compare_cpu_decoing_time}
\end{figure*}

In this section, we only focus on the consuming time of translating the entire MT03 test dataset. Under the two conditions, we calculated the times spent on translating the entire test dataset for different beam sizes, then draw the curves in Figure \ref{fig:compare_gpu_decoing_time} and \ref{fig:compare_cpu_decoing_time}. From the Figure \ref{fig:gpu-nist03-osn-time} and \ref{fig:gpu-nist03-sn-time}, we could observe that accelerated cube pruning algorithm speeds up the decoding by about $3.8\times$ on GPUs when the beam size is set to $40$. Figure \ref{fig:cpu-nist03-osn-time} and \ref{fig:cpu-nist03-sn-time} show that the accelerated cube pruning algorithm speeds up the decoding by about $4.2\times$ on CPU server with the beam size $40$.

\section{Related Work} 	\label{sec5}

Recently, lots of works devoted to improve the efficiency of the NMT decoder. Some researchers employed the way of decreasing the target vocabulary size. \citet{jean2014using} improved the decoding efficiency even with the model using a very large target vocabulary but selecting only a small subset of the whole target vocabulary. Based on the work of \citet{jean2014using}, \citet{mi2016vocabulary} introduced sentence-level and batch-level vocabularies as a very small subset of the full output vocabulary, then predicted target words only on this small vocabulary, in this way, they only lost $0.1$ BLEU points, but reduced target vocabulary substantially.

Some other researchers tried to raise the efficiency of decoding from other perspectives. \citet{wu2016google} introduced a coverage penalty $\alpha$ and length normalization $\beta$ into beam search decoder to prune hypotheses and sped up the search process by $30\%$$\sim$$40\%$ when running on CPUs. \citet{hu2015improved} used a priority queue to choose the best hypothesis for the next search step, which drastically reduced search space.

Inspired by the works of \citet{mi2016vocabulary} and \citet{huang2007forest}, we consider pruning hypothesis in NMT decoding by using cube pruning algorithm, but unlike traditional SMT decoding where dynamic programming was used to merge equivalent states (e.g., if we use phrase-based decoding with trigram language model, we can merge states with same source-side coverage vector and same previous two target words). However, this is not appropriate for current NMT decoding, since the embedding of the previous target word is used as one input of the calculation unit of each step in the decoding process, we could group equivalence classes containing the same previous target word together.

\section{Conclusions}
\label{sec6}

We extended cube pruning algorithm into the decoder of the attention-based NMT. For each step in beam search, we grouped similar candidates in previous beam into one or more equivalence class(es), and bad hypotheses were pruned out. We started searching from the upper-left corner in each equivalence class and spread out until enough candidates were generated. Evaluations show that, compared with naive beam search, our method could improve the search quality and efficiency to a large extent, accelerating the NMT decoder by $3.3\times$ and $3.5\times$ on GPUs and CPUs, respectively. Also, the translation precision could be the same or even better in both situations. Besides, self-normalization is verified to be helpful to accelerate  cube pruning even further.

\section*{Acknowledgements}
We thank the three anonymous reviewers for their comments, Kai Zhao and Haitao Mi for uggestions. 
This work is supported in part by NSF IIS-1817231 \& IIS-1656051, and is also supported in part by National Natural Science Foundation of China (No.~61472428 \& No.~61662077).

\bibliography{emnlp2018}
\bibliographystyle{acl_natbib_nourl}

\end{document}